\documentclass[10pt,twocolumn]{article}
\usepackage{cvpr}

\AtBeginDocument{%
  \providecommand\BibTeX{{%
    \normalfont B\kern-0.5em{\scshape i\kern-0.25em b}\kern-0.8em\TeX}}}
    
\usepackage{graphicx}
\usepackage{amsmath}
\usepackage{amssymb}
\usepackage{booktabs}

\usepackage{multirow}
\usepackage{mathrsfs}
\usepackage{array}
\usepackage{color}
\usepackage{ifpdf}
\usepackage{algorithmic}
\usepackage{algorithm}
\usepackage{graphicx}
\usepackage{xcolor}
\usepackage{makecell}
\usepackage[utf8]{inputenc}

\usepackage[pagebackref,breaklinks,colorlinks,bookmarks=false]{hyperref}

\usepackage[capitalize]{cleveref}
\crefname{section}{Sec.}{Secs.}
\Crefname{section}{Section}{Sections}
\Crefname{table}{Table}{Tables}
\crefname{table}{Tab.}{Tabs.}

\def\confName{CVPR}
\def\confYear{2023}

\title{ReGeneration Learning of Diffusion Models with Rich Prompts for Zero-Shot Image Translation}
\begin{document}

\author{Yupei Lin$^1$\hspace{.3cm} Sen Zhang$^2$\hspace{.3cm} Xiaojun Yang$^1$\hspace{.3cm} Xiao Wang$^3$\hspace{.3cm}  Yukai Shi$^1$ \thanks{Corresponding author: ykshi@gdut.edu.cn}\\[0.10cm]
$^1$Guangdong University of Technology\hspace{.4cm} $^2$The University of Sydney\hspace{.4cm} $^3$Anhui University\\[0.11cm]
}

\maketitle
\begin{abstract}
\vspace{-0.1cm}
    Large-scale text-to-image models have demonstrated amazing ability to synthesize diverse and high-fidelity images. However, these models are often violated by several limitations. Firstly, they require the user to provide precise and contextually relevant descriptions for the desired image modifications. Secondly, current models can impose significant changes to the original image content during the editing process. In this paper, we explore ReGeneration learning in an image-to-image Diffusion model (ReDiffuser), that preserves the content of the original image without human prompting and the requisite editing direction is automatically discovered within the text embedding space. To ensure consistent preservation of the shape during image editing, we propose cross-attention guidance based on regeneration learning. This novel approach allows for enhanced expression of the target domain features while preserving the original shape of the image. In addition, we introduce a cooperative update strategy, which allows for efficient preservation of the original shape of an image, thereby improving the quality and consistency of shape preservation throughout the editing process. Our proposed method leverages an existing pre-trained text-image diffusion model without any additional training. Extensive experiments show that the proposed method outperforms existing work in both real and synthetic image editing.The project page is available at  \url{https://yupeilin2388.github.io/publication/ReDiffuser}
\vspace{-0.4cm}
\end{abstract}

\maketitle
\section{Introduction}

 Recent text-image diffusion models have demonstrated notable generalization capabilities, particularly when trained on large-scale text-image pair datasets such as LAION-5B. These large-scale text-image diffusion models are capable of generating high-quality images with diverse content~\cite{brooks2022instructpix2pix,meng2021sdedit,mokady2022null,parmar2023zero,wu2022unifying}, rich texture, and editable semantics. However, these diffusion models still have drawbacks in zero-shot image-to-image translation. Firstly, they require an accurate description of a given image, which can be difficult to express. Furthermore, current text-to-image diffusion models show a tendency to alter the shape of the source image in image-to-image conversion tasks, resulting in deviations from its original appearance. SDEdit~\cite{meng2021sdedit} employs an initial step of introducing a minute level of noise to the input, followed by subsequent denoising that is conditioned on the target prompt. The prompt-to-prompt~\cite{hertz2022prompt} approach leverages the attention map of the original image to guide the spatial structure of the target prompt. Instruct Pix2Pix~\cite{brooks2022instructpix2pix} employs a diffusion model of the image condition to meticulously adhere to the user's prompt. Although these methods are applicable for image editing tasks, they may not effectively preserve shape consistency between the original and edited images. Ensuring shape consistency throughout the editing process remains a challenge.

\begin{figure*}[h]
\centering
\includegraphics[width=0.98\linewidth]{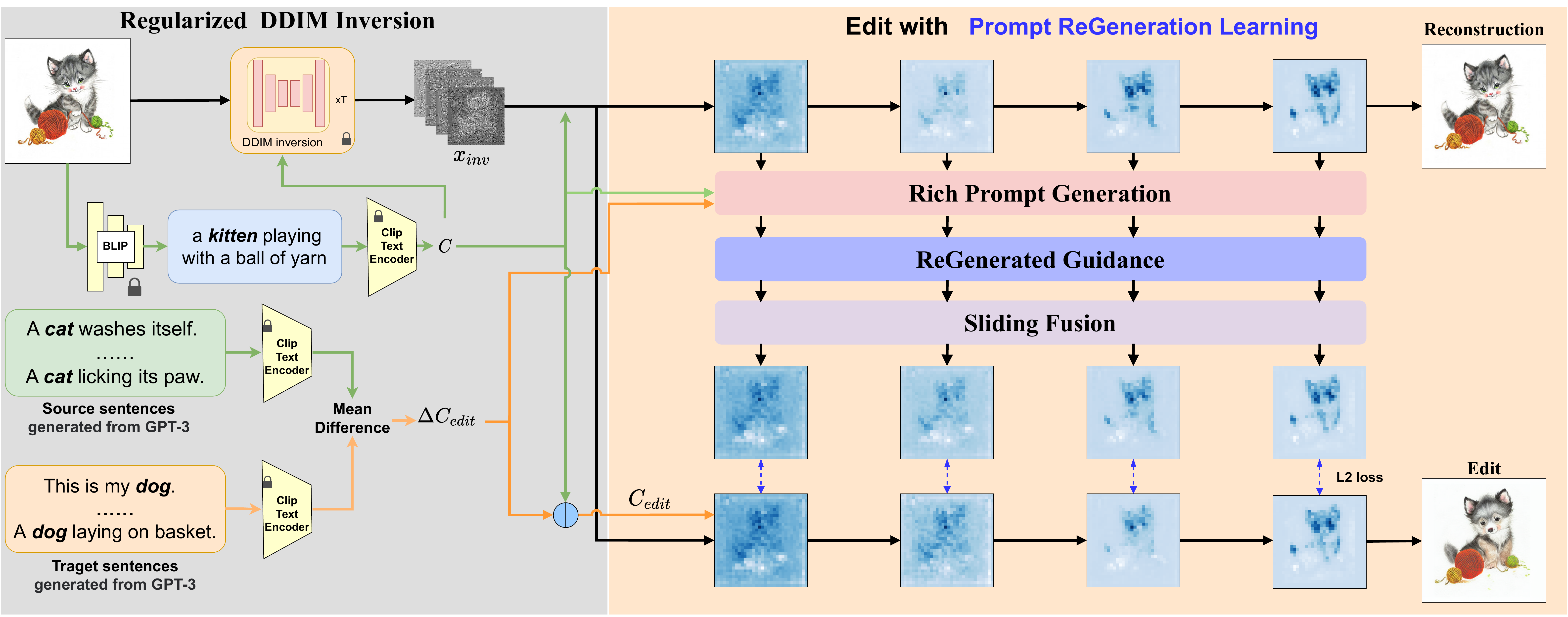}
\caption{The overview of ReDiffuser. We employed the BLIP~\cite{li2022blip} and CLIP~\cite{radford2021learning} text encoder to calculate the embedding C of the original text, which was used for DDIM inversion to obtain an inversion noise. We then use GPT-3 to generate several sentences that involved source domain (e.g., cat) and target domain (e.g., dog) to calculate the mean difference between them denoted as $\bigtriangleup c_{\text{edit}}$. Next, we use the original text embedding to denoise the inversion noise to obtain cross-attention maps. We input these cross-attention maps into the prompt regeneration learning, and the regenerated result will be used as the reference for image editing (second row). Finally, we use the edited text embedding $c_{\text{edit}}$ to denoise and encourage the cross-attention map to match the reference cross-attention map (third line). }
\label{fig:farmework}
\end{figure*}

The Pix2Pix-Zero method~\cite{parmar2023zero}, which has been recently proposed, encompasses three stages for image editing: inversion, reconstruction, and editing. First, the image is inverted into noise using DDIM. Then, during the reconstruction phase, text features are extracted from the original image and used as conditional text input along with the noisy image in a diffusion model to reconstruct the image. Finally, in the editing phase, the difference between the features in the source and target domains is calculated and used as additional conditional text input to guide the direction of image editing. The shape consistency during the editing process is maintained through cross-attention, which is influenced by the reconstruction process. Nevertheless, this approach has certain limitations. The utilization of cross-attention in the source domain may impede the representation of features in the target domain, resulting in images that retain their shape but fail to capture the features of the target domain. Additionally, shape consistency may not be well-maintained in cases where the cross-attention in the source and target domains exhibit significant differences.
In this paper, we propose ReDiffuser, a straightforward and efficient diffusion model for zero-shot image translation. Unlike previous image editing methods, our approach eliminates the need to train on each target prompt and does not require the user to give a mask as an editing prompt. Inspired by the limitations of previous methods in preserving shape consistency during image editing, as well as in feature representation for editing directions, we present two novel innovations to tackle these challenges:
As illustrated in Fig.~\ref{fig:farmework}, our method did not solely rely on the attention map in each step of the reconstruction phase. Instead, we utilized ReGeneration learning to generate cross-attention maps with rich prompts. We then regenerate these cross-attention maps by sliding fusion and guide the subsequent editing phase of the DDIM with the regenerated results.
This prompt regeneration learning effectively preserves the original structure, ensuring that it remains unchanged throughout the image editing process. Furthermore, we propose a cooperative update strategy to help the generative model to learn representation from the regenerated soft guidance.
With the support of our novel designs, we can edit images directly using pre-trained text-image models and maintain the appearance.
 \begin{figure}[t]
\centering
\includegraphics[width=0.69\linewidth]{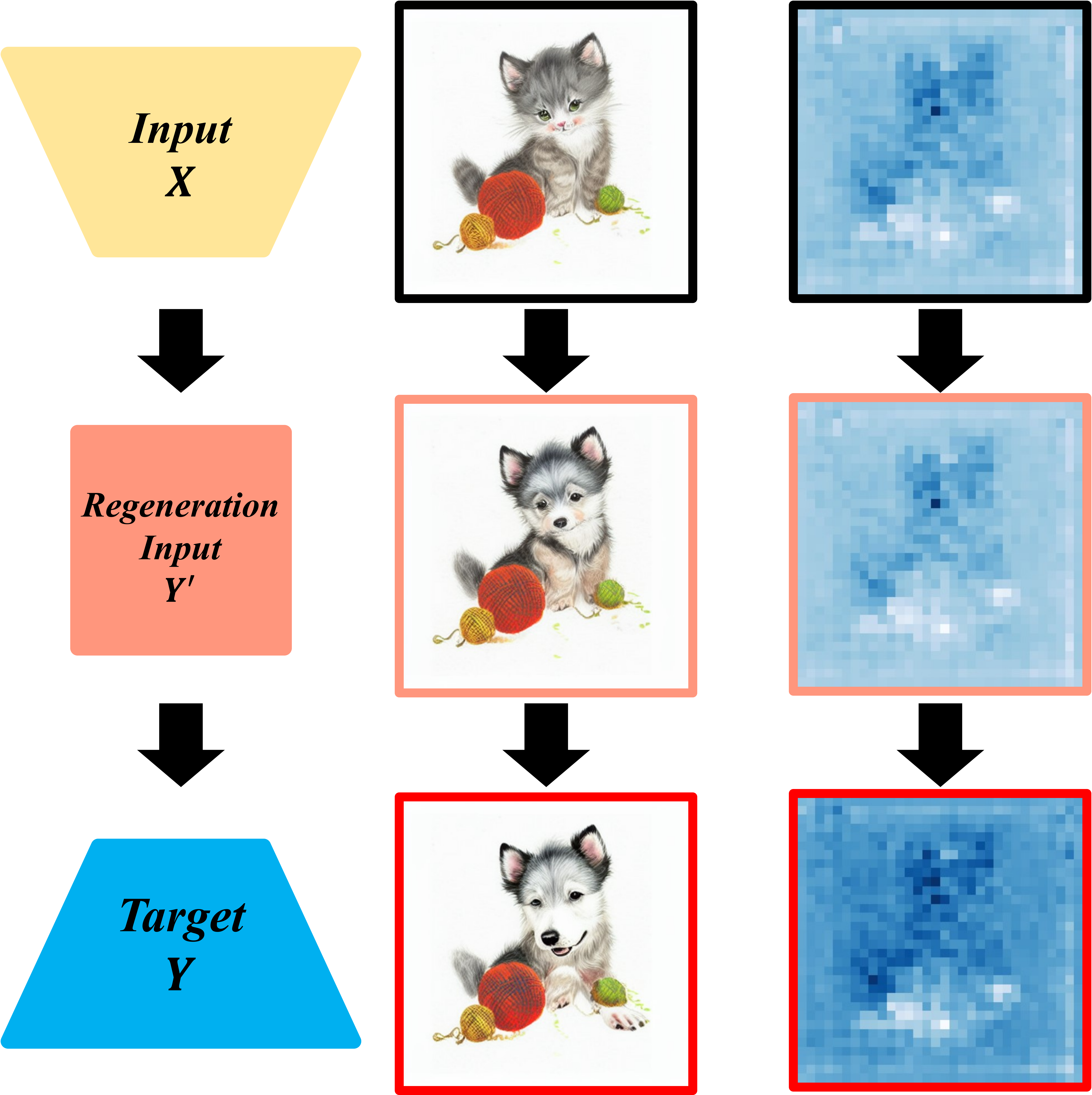}
\caption{ReGeneration learning. In contrast to previous work, regenerative learning involves learning from the input $X$ to a more concise representation $Y'$, and subsequently learning from $Y'$ to generate the final output $Y$.}
\label{fig:RL}
\end{figure}

Our contributions are summarized as follows:
\vspace{-0.2cm}
\begin{itemize}
    \item Given the limitations of previous diffusion-edit methods in the editing phase, we applied ReGeneration Learning to the reconstruction phase and obtained rich prompt cross-attention maps that contained contextual features of the source and target domain features. 
    \vspace{-0.2cm}
    \item Rather than simply fusing the cross-attention maps of rich prompts at the current state, we propose an approach called Sliding Fusion which utilizes a sliding window to fuse cross-attention information of rich prompts across adjacent time steps. This approach further helps diffusion model to better represent the features associated with the editing direction, while still preserving the shape of the image.  
    \vspace{-0.2cm}
    \item We introduce a cooperative update strategy to help the diffusion model to learn effective representation from the regenerated soft guidance, further enhancing the overall quality of the edited images.
    \vspace{-0.2cm}
    \item We collected three challenging datasets on cat, horse, and sketch, and evaluated recent diffusion-based image editing works and our approach on four image editing tasks (Cat $\rightarrow$ Dog-Free, Cat with glasses, Horse $\rightarrow$ Zebra-Free and Sketch$\rightarrow$ Oil-Free). Experimental results show that our method has a good generation effect and structure retention. Furthermore, each subassembly in ReDiffuser is carefully analysed by ablation experiments to verify its effectiveness. 
    \vspace{-0.2cm}
\end{itemize}

 \section{Related Work}

 \textbf{Text-to-Image Diffusion model.} Current Text-to-Image diffusion models greatly improve image quality and diversity by training on large text-to-image dataset, which in this way enables the generation of high-quality images from text~\cite{yu2022scaling,ding2021cogview}. DALL$\cdot$E2~\cite{ramesh2022hierarchical} encodes the input text description as a text embedding via CLIP~\cite{radford2021learning}, then maps it to the image encoding of its corresponding image via a diffusion model, and finally decodes the image encoding into the generated image using a decoder. Low-resolution latent space diffusion (LDM)~\cite{rombach2022high} effectively mitigates the computational burden by diffusing over a lower-resolution latent space. This approach enables the training of text-to-image diffusion models on large-scale datasets such as LAION-5B~\cite{schuhmann2022laion}, rendering it the most widely utilized image-to-text diffusion model.
 
 \textbf{Image editing with diffusion models.} Image editing often requires that the appearance of the image remains unchanged before and after editing, and datasets used for this task are often unpaired. In previous approaches, image editing tasks were commonly accomplished using image-to-image translation methods ~\cite{zhu2017unpaired,park2020contrastive,lin2022exploring,zheng2021spatially} in Generative Adversarial Networks~\cite{goodfellow2014generative}. Several recent works have utilized diffusion models for image editing. SDEdit~\cite{meng2021sdedit} enhances image realism by introducing noise into the input image and subsequently denoising it with a pretrained diffusion model. Imagic~\cite{kawar2022imagic} uses a pretrained text-to-image diffusion model and fine-tunes the model to capture the appearance of specific image features. Prompt-to-Prompt~\cite{hertz2022prompt} allows text prompts to be replaced, refined and re-weighted for editing to manipulate the edited image. However, these approaches still have challenges in maintaining shape and expressing the editing direction.

\begin{figure*}[h]
\centering
\includegraphics[width=0.89\linewidth]{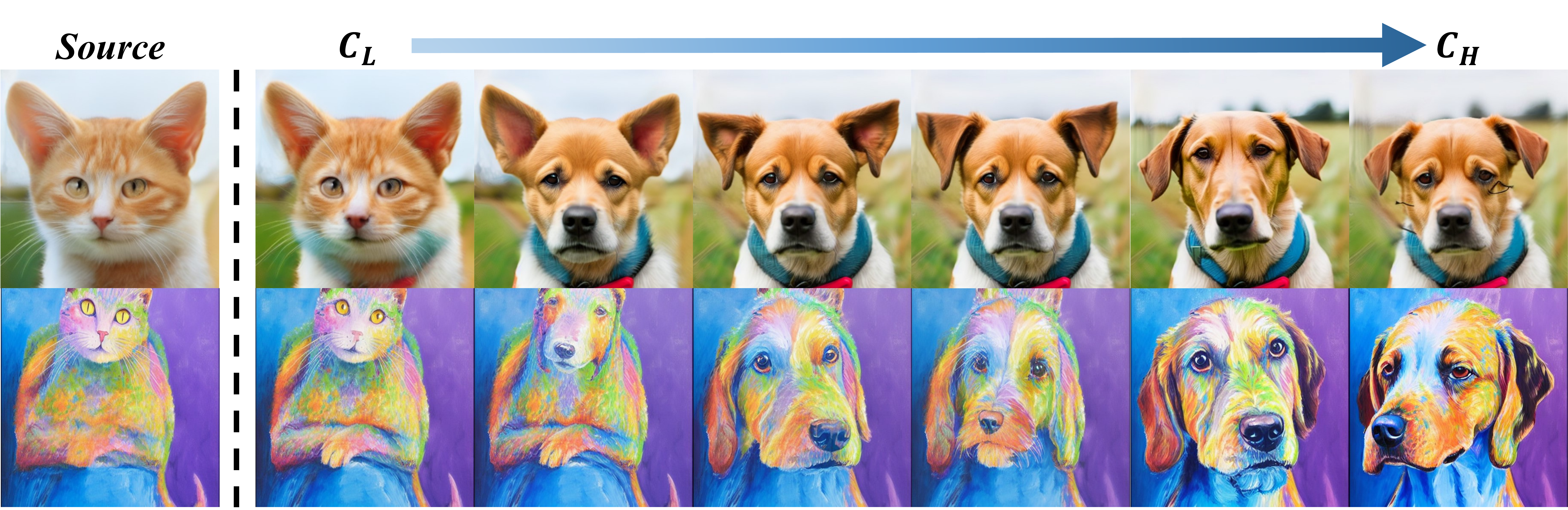}
\caption{Generated results with different hardness of prompts. For the cat to dog task, we utilized an initial prompt $c$ in combination with a set of $\bigtriangleup c_{\text{edit}}$ with different weights as the editing direction to edit image. Upon observing the generated results from left to right, we noticed that the images generated by combining the initial prompt $c$ with the prompt $\bigtriangleup c_{\text{edit}}$ of larger weights exhibit more of the desired editing direction, i.e., the dog features. This motivates us to use those rich prompts in diffusion model for a high-fidelity image translation. }
\label{fig:RPI}
\end{figure*}
 
 \textbf{ReGeneration learning.} Regeneration learning~\cite{tan2023regeneration} is an emerging paradigm in data generation, designed to address the challenges faced by traditional conditional data generation methods. In conditional approaches, the target data $Y$ (such as text, speech, music, images, etc.) often exhibits high dimensionality and complexity, and may contain information not present in the source data X, making direct mapping from $X$ to $Y$ difficult to learn effectively and efficiently. The core idea of regeneration learning is to represent the target data $Y$ as an intermediate layer denoted as $Y'$, and then learn the mappings from $X$ to $Y'$ and from$Y'$ to $Y$, respectively. This method reduces the difficulty of the data generation task, as the mapping from $X$ to $Y'$ and from $Y'$ to $Y$ is simpler compared to the direct mapping from $X$ to $Y$.

\section{Method}


Our goal is to achieve zero-shot text-driven image-to-image translation (I2I) for style, attributes, and shape. Unlike previous approaches~\cite{nichol2021glide,avrahami2022blended,couairon2022diffedit} that require optimizing the target with a given mask, our approach only requires an input image and an editing direction to achieve I2I. 

In this paper, we utilize a pre-trained stable diffusion model, which takes the input image $x\in R^{X\times X\times 3}$ and encodes it as a latent code $x_0\in R^{S\times S\times 4}$ using VQ-VAE~\cite{van2017neural}, where $X$ is the image size of 512 and $S=64$ is the downsampled latent image size. The inversion and editing operations described in this section are performed in the latent image space. During the DDIM inversion procedure, an initial text prompt is required as conditional text input. In this paper, we utilized BLIP~\cite{li2022blip} to generate the initial text describing the input image, and input the text into CLIP to obtain the initial prompt $c$.

\subsection{Preliminaries}
\textbf{Diffusion Model.}
The diffusion model~\cite{ho2020denoising} learns the inverse process of the diffusion process, which can reconstruct the distribution of the data. The diffusion process is modeled as a Markov process with $x_t$ denoting the random variable at the $t$-th time step.
\begin{equation}
\boldsymbol{x}_{t} \sim \mathcal{N}\left(\sqrt{\alpha_{t}} \boldsymbol{x}_{t-1}, \left(1-\alpha_{t}\right) \boldsymbol{I}\right)
\end{equation}
where $\alpha_t$ is a fixed coefficient that determines the noise schedule. The above definition leads to a simple approximation of $p(x_t|x_0)$ in the form:

\begin{equation}
x_{t}=\sqrt{\bar{\alpha}_{t}}x_{0}+\sqrt{1-\bar{\alpha}_{t+1}} \epsilon
\label{eq:2}
\end{equation}

where $\varepsilon$ denotes noise
\begin{equation}
\bar{\alpha}_{t} = \prod_{s=1}^{t} a_s,\epsilon N(0,I)
\end{equation}
This further allows to efficiently sample arbitrary $x_t$ during the training process.

\textbf{Latent Diffusion Model.} The Latent Diffusion Model (LDM) utilizes the latent space of an autoencoder~\cite{kingma2013auto} for diffusion and denoising. The RGB image $y$ is first compressed into a low-resolution latent representation $x$ through the encoder $E$, and it can be reconstructed back into the image by the decoder $D$ such that $D(x) \approx y$. The training objective of the LDM can be expressed using the reparameterization trick as:
\begin{equation}
    \min _{\theta} E_{x_{0}, \varepsilon \sim N(0, I), t \sim \text { Uniform }(1, T)}\left \|\varepsilon-\varepsilon_{\theta}\left(x_{t}, t, c\right) \right \|_{2}^{2}
\end{equation}
where c is the embedding of the conditional text prompt, and $x_t$ is the noise sample at time step t of $X_0$, computed by equation~\ref{eq:2}. $\varepsilon_{\theta}$ denotes the U-Net network used to remove the noise.

\textbf{DDIM Inversion.} We desire the inversion process to generate a noise map $x_{inv}$ that accurately reconstructs our latent code $x_0$ after t timesamples. However, in DDPM, the sampling and inversion processes~\cite{song2020denoising,dhariwal2021diffusion} are stochastic, and thus the resulting inversion may not faithfully reconstruct the input. As a solution, we adopt deterministic DDIM for both the sampling and inversion processes.

In the sample phase, the random noise $x_t$ in the sequence of time steps t: T$\rightarrow$1 is gradually converted to a clean latent code $x_0$ by a deterministic DDIM sample.
\begin{equation}
    x_{t-1}=\sqrt{\bar{\alpha}_{t-1}} f_{\theta}\left(x_{t}, t, c\right)+\sqrt{1-\bar{\alpha}_{t-1}} \epsilon_{\theta}\left(x_{t}, t, c\right)
\end{equation}
where $f_\theta(x_t, t, c)$ predicts the final denoising latent code $x_0$, denoted as:
\begin{equation}
f_\theta(x_t,t,c) = \dfrac{x_t-\sqrt{(1-\bar{a})}\epsilon_{\theta}\left(x_{t}, t, c\right)}{\sqrt{\bar{a_t}}}
\end{equation}
In the inversion phase, a clean latent code $x_0$ is inverted into a nosie latent ${x_t}$ based on the ODE limit analysis of the diffusion process with time step $t : 1 \rightarrow T$:
\begin{equation}
{x}_{t}=\sqrt{\bar{\alpha}_{t}} f_{\theta}\left({x}_{t-1}, t-1, c\right)+\sqrt{1-\bar{\alpha}_{t}} \epsilon_{\theta}\left({x}_{t-1}, t- 1, c\right)
\end{equation}
By adding noise gradually, it can make the input $x_0$ gradually become noise by DDIM, and finally get the latent code $x_{inv}$.


\textbf{Noise regularization.} We assume that each time step is well regularized in the diffusion context, but relying on multiple iterations causes the noise of the intermediate time step to be out of the Gaussian distribution. So we guide the inversion process with an autocorrelation loss, this loss consists of a pairwise term $\mathcal{L}_{pair}$ and a KL divergence term $\mathcal{L}_{KL}$ at individual pixel location. Since it is very costly to sample all pairs of locations frequently, a feature pyramid is constructed for this purpose. In this pyramid, the predicted noise map $\epsilon_{\theta}$ is utilized as the initial noise $\eta^0 \in \mathbb{R}^{64 \times 64 \times 4}$, and each subsequent noise map is subjected to 2$\times$ 2 pooling with an averaging filter. The pooling process stops at a feature size of 8 $\times$ 8 and creates four noise maps, namely $\eta^0, \eta^1, \eta^2, \eta^3$. These noise maps are used for the calculation of $\mathcal{L}_{pair}$, which is as follows:

\begin{equation}
    \mathcal{L}_{pair} = \sum_p{\dfrac{1}{S^2_p}}\sum_{\delta }^{S_p-1}\sum_{x,y,c}\eta^p_{x,y,c}(\eta^p_{x-\delta,y,c}+\eta^p_{x,y-\delta,c}),
\end{equation}

where $p$ is the level of the pyramid, $S_p$ denotes the noise map sizes, $\sigma$ denotes the possible offsets, $\eta^p_{ x,y,c} \in R$ indexes into a spatial location, circular index, and channel index. In addition, we also use $L_{KL}$, and the final autocorrelation regularization loss we use in the inversion phase is $\mathcal{L}_{auto} = \mathcal{L}_{pair} + \lambda \mathcal{L}_{KL}$, where $\lambda$ balances these two losses and we improve the effect of inversion by this regularization loss.

\begin{figure*}[ht]
\centering
\includegraphics[width=0.7\linewidth]{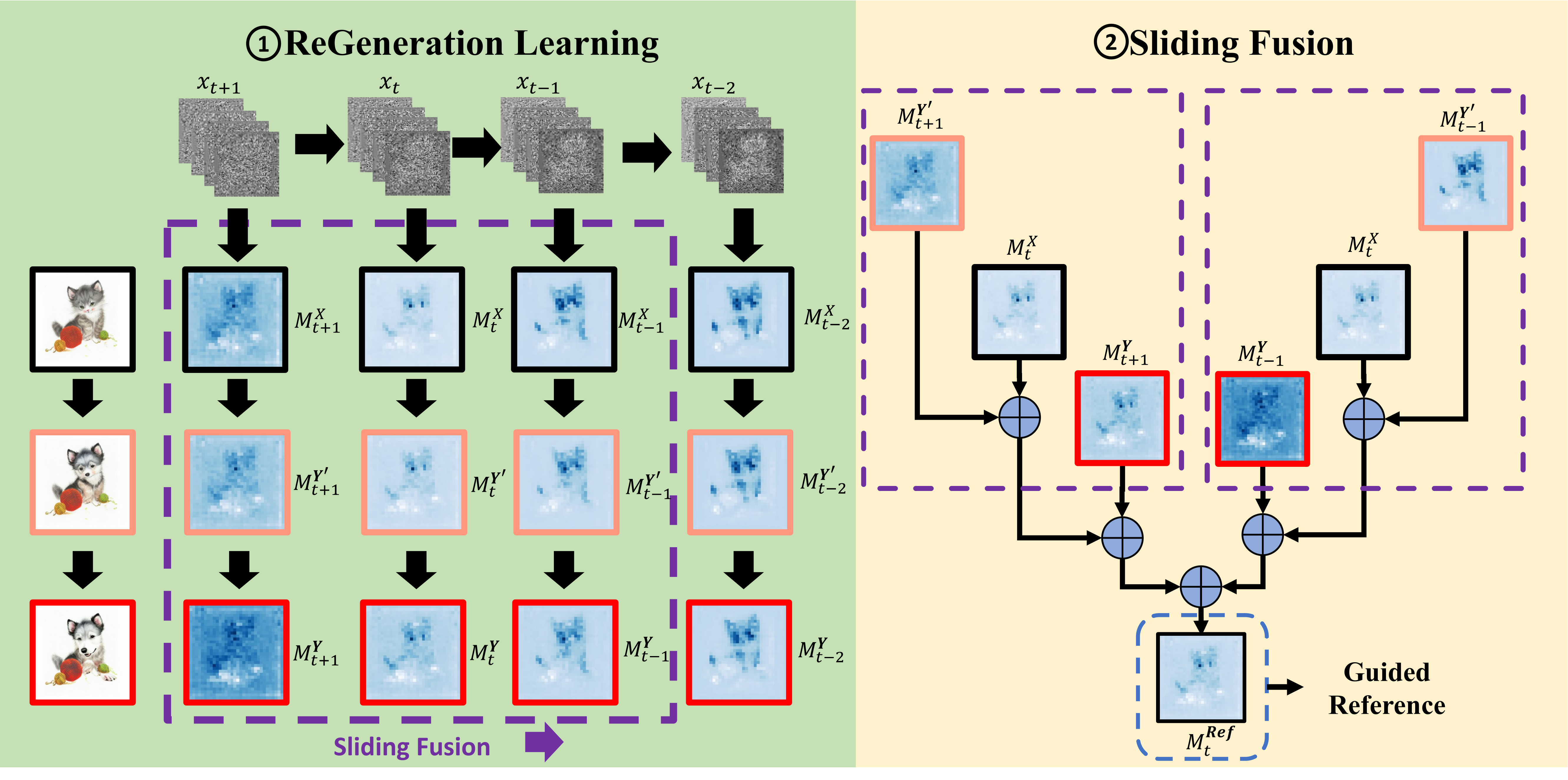}
\caption{ReGeneration Learning with rich prompts. We identified rich prompts that corresponded with the ReGeneration Learning method to generate three cross-attention maps referred to as $M^{X}$, $M^{Y'}$, and $M^{Y}$ at each time step of the reconstruction phase. These cross-attention maps are aligned with the inputs, labels, and targets of the regeneration learning. Next, we used sliding prompt fusion to merge the regenerated input with the labels and targets from the preceding time steps to obtain fused results. This fusion result was used as a guide reference for the editing phase.}
\label{fig:BPF}
\end{figure*}
\subsection{Typical Prompt Generation}
The recent approach necessitates a particular sentence as a prompt to regulate image generation, but it is often challenging for users to provide an exact prompt. We expect that the prompt would only require the user to specify the words that differentiate between the source and target domains to compute the embedding $\bigtriangleup c_{\text{edit}}$ of our prompt. Hence, we propose prompt generation. By inputting the source domain word and the target domain word into GPT-3~\cite{brown2020language}, a substantial number of sentences related to each domain are generated. These generated sentences are then fed into CLIP Text Encoder to calculate the average disparity between the two domains, the detailed calculations are as follows:
\begin{equation}
    \bigtriangleup c_{edit} = \dfrac{1}{N}\sum_{i=1}^N(\text{CLIP}_{\text{text}}(\hat{s}_{i})-\text{CLIP}_{\text{text}}({s}_{i}))
\end{equation}
 where N is the number of sentences generated by GPT-3, $\text{CLIP}_{\text{text}}$ denotes the CLIP text encoder, $s_i$ and $\hat{s_i}$ denote the i-th source domain sentence and the i-th target domain sentence generated by GPT-3, respectively.
In contrast to other methods, this approach offers the advantage of pre-calculating the direction, obviating the need for repetitive calculations with each usage. Furthermore, to apply the edit, we add the pre-calculated edit direction $\bigtriangleup c_{\text{edit}}$ to the initial prompt $c$ to get $c_{\text{edit}}$ as the edit direction.

\subsection{ReGeneration Learning in Diffusion Models}

\textbf{Cross-attention in Diffusion Models.} The cross-attention map in the LDM model, which captures the interplay between feature maps and conditional text features, demonstrates a desirable locality property in the well-trained text-to-image diffusion model. This advantageous property allows us to precisely specify the target features that necessitate modification by using the cross-attention map as a reliable guide. The cross-attention map in the LDM model is computed as follows:
\begin{equation}
    \left\{\begin{array}{l}
\mathbf{Q}=\mathbf{W}_{Q} \cdot \varphi \left( x_t \right) ; \mathbf{K}=\mathbf{W}_{K} \cdot \mathbf{c} ; \mathbf{V}=\mathbf{W}_{V} \cdot \mathbf{c} \\
\mathbf{M}=\operatorname{Softmax}\left(\frac{Q K^{T}}{\sqrt{d}}\right) \\
\operatorname{Attention}(\mathbf{Q}, \mathbf{K}, \mathbf{V})=\mathbf{M} \cdot \mathbf{V}
\end{array}\right.
\end{equation}
where $W_Q$, $W_K$, $W_V$ are learnable projections, $\varphi \left( x_t \right)$ is the intermediate space features of the U-Net $\epsilon_{\theta}$, $\mathbf{c}$ denotes text embedding, and d is the dimension of keys $\mathbf{K}$ and queries $\mathbf{Q}$~\cite{vaswani2017attention}. 
It can be observed that the cross-attention map is closely associated with the structure of the image, a single entry of the mask $M_{i,j}$ represents the contribution of the j-th text token to the i-th spatial location. Furthermore, it's important to note that the cross-attention mask is temporally variant, resulting in a distinct attention mask $M_t$ at each time step $t$. Thus, we can leverage the cross-attention map from the reconstruction phase to provide positional information for the editing phase. 

\textbf{Rich Prompt Generation.} 
Previous research utilizes the cross-attention map of the original image as a hard constraint~\cite{hertz2022prompt} or soft guide~\cite{parmar2023zero} to regulate the appearance of the edited image. However, using the cross-attention map from the reconstruction phase as a reference may restrict the representation of features in the editing direction. Since the cross-attention map in the reconstruction process merely contains the features of the original image, which are not identical to the features of the target domain. As shown in Fig.~\ref{fig:RPI}, we make an experiment by combining the reconstruction prompt $c$ with $\bigtriangleup c_{\text{edit}}$ of different values, gradually increasing from the smaller value prompt $C_L$ to $C_H$, and using these prompts as edit directions to generate results. It can be observed that there is a gradual shift from left to right in the generated results, progressing from cat features to increasingly dog-like features. This indicates that prompts with different hardness contain different degrees of transitional features from cat to dog. We then propose an assumption: using a single reconstructed prompt as the reference would constrain the expression of the target domain, a diverse range of prompts will help to learn rich features? We refer to these diverse prompts as rich prompts in our ReGeneration procedure. 

\textbf{ReGeneration Learning.} We used the different hardness of the prompt $c$, $c_m$, and $c_h$ as edit directions to generate the images. As shown in Fig.~\ref{fig:RL}, it can be seen that the generated results with rich prompts as the edit direction represent the cat feature, the transition feature between cat and dog, and the dog feature, respectively, corresponding exactly to the input $X$, the label $Y{'}$ and the target $Y$ for regeneration learning. Similarly, we can use these rich prompts as editing directions with the current state of the reconstruction process $x_t$ input to U-Net, resulting in three attention maps $M^{X}_t$, $M^{Y^{'}}_t$ and $M^{Y}_t$, these attention maps correspond to the input, label, and target of the ReGeneration procedure, respectively. As shown in Step 1 of Fig.~\ref{fig:BPF}, we apply the rich prompts with ReGeneration learning into the reconstruction process. We fuse the attention maps $M^{X}$, $M^{Y^{'}}$, and $M^{Y}$ via a sliding window, which serves as reference during the editing phase. 

\textbf{Sliding Fusion.} We propose sliding fusion, a sliding window with stride = 1 and window size = 3, which makes full use of the temporal contextual relationships in the cross-attention map. As shown in Step 2 of Fig.~\ref{fig:BPF}, we combine the cross-attention maps of the ReGeneration labels and targets of adjacent time steps with the cross-attention map of the ReGeneration input of the current state. This fused attention map retains the appearance features of the original image while fusing the target domain representation across adjacent time steps, resulting in a smooth transition effect. We refer to the fused attention map as $M^{\text{Ref}}$, which serves as a guidance for the editing phase.

 \subsection{Editing DDIM} 
During the editing phase, suppose we only apply the edit direction prompt $c_{\text{edit}}$ for DDIM editing, the results will exhibit the intended edit direction features, but fail to maintain the original shape. To guide the editing process, we utilized the $M^{Ref}$ acquired through the sliding prompt fusion as a reference to softly guide the editing phase. By employing $x_t$ as the gradient to align with the reference $M^{\text{Ref}}t$, we aim to minimize the cross-attention loss $\mathcal{L}_{xa}$:

\begin{equation}
  \mathcal{L}_{xa} = \left \| M^{edit}_t - M^{Ref}_t \right \|_2  
\end{equation}
where $M^{edit}$ denotes the cross-attention map for the editing phase. This loss encourages our $M^{edit}_t$ keep the structure consistent with reference map $M^{Ref}_t$.

\begin{figure}[t]
\centering
\includegraphics[width=0.6\linewidth]{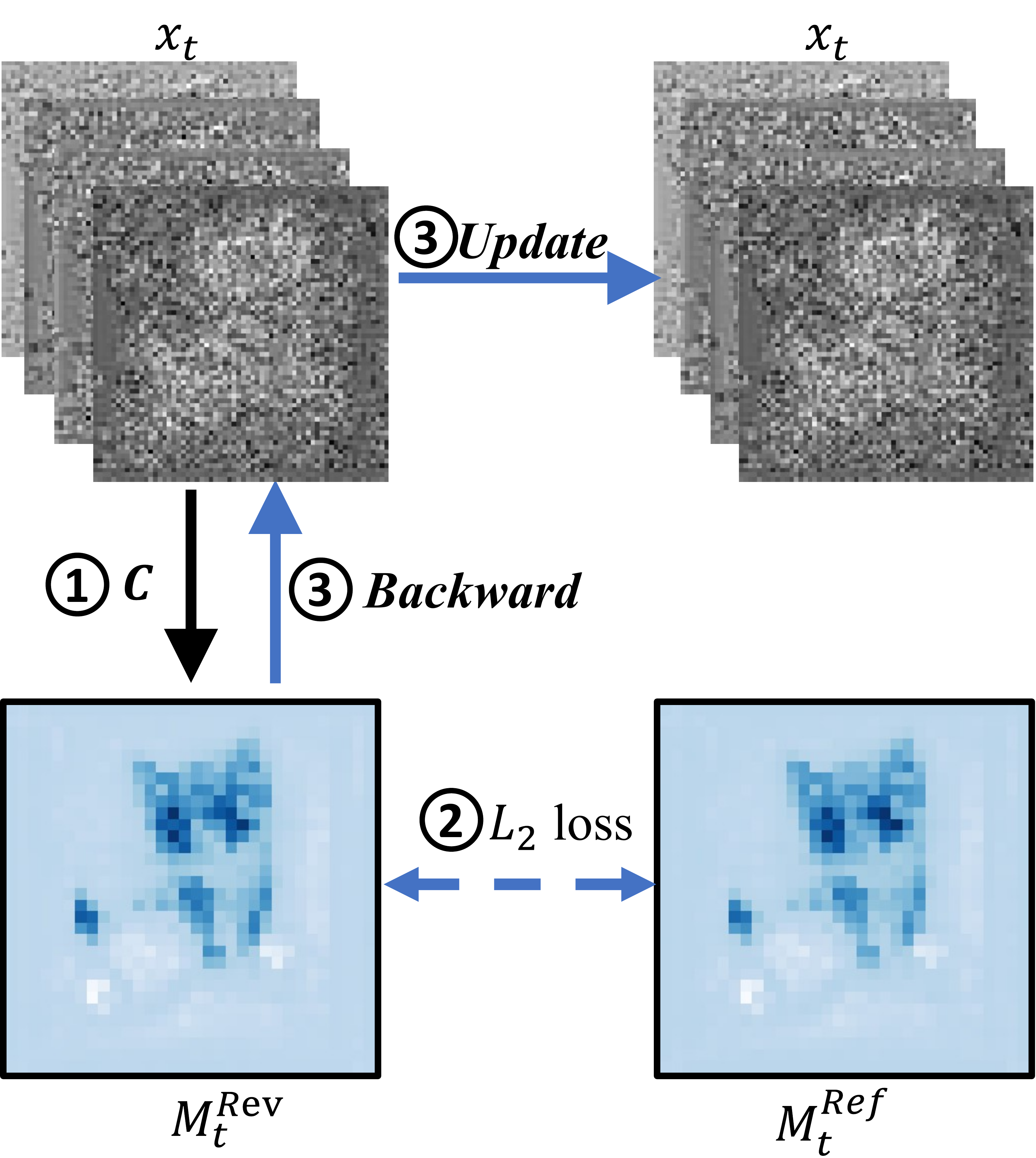}
\caption{Cooperative Update among regenerated guidance for DDIM editing. Our Cooperative Update strategy help the generative model to learn effective representation from a soft guidance.}
\label{fig:meta_cycle}
\end{figure}
\textbf{Cooperative Update.} In the process of generation, we found that using the $l_2$ guidance alone was not sufficient to maintain the shape. 
We propose a Cooperative Update strategy to further maintain the appearance, as shown in Fig.~\ref{fig:meta_cycle}, where we perform the following three steps in the editing process:
\begin{itemize}
    \item[1)] \textbf{Reverse Generation.} At time step t, the generated sample $x_t$, which is sampled in the direction of the source to the target domain, contains features in the target domain. We input this $x_t$, along with the prompt from the reconstruction phase, into U-Net to perform a one-step reverse generative operation from target to source. The resulting cross-attention map is denoted as $M^{\text{Rev}}_t$.
    \item[2)] \textbf{Calculated differences.} To ensure consistency before and after image editing, we aim for $M^{\text{Rev}}_t$ to be similar to $M^{\text{Ref}}_t$. The difference between them can be quantified by calculating the $L_2$ distance:
    \begin{equation}
    \mathcal{L}_{\text{rev}} =  \lVert M^{\text{Rev}}_t - M^{\text{Ref}}_t \rVert_2
    \end{equation}
    \item[3)] \textbf{Cooperative Update.} Update $x_t$ at the current moment by back-propagating the gradient in order to minimize the loss:
    \begin{equation}
    x_t = x_t -\lambda_{rev}\bigtriangleup{x_t}(\left \| M^{Rev}_t - M^{Ref}_t \right \|_2)
    \end{equation}
\end{itemize}

\begin{figure*}[t]
\centering
\includegraphics[width=0.89\linewidth]{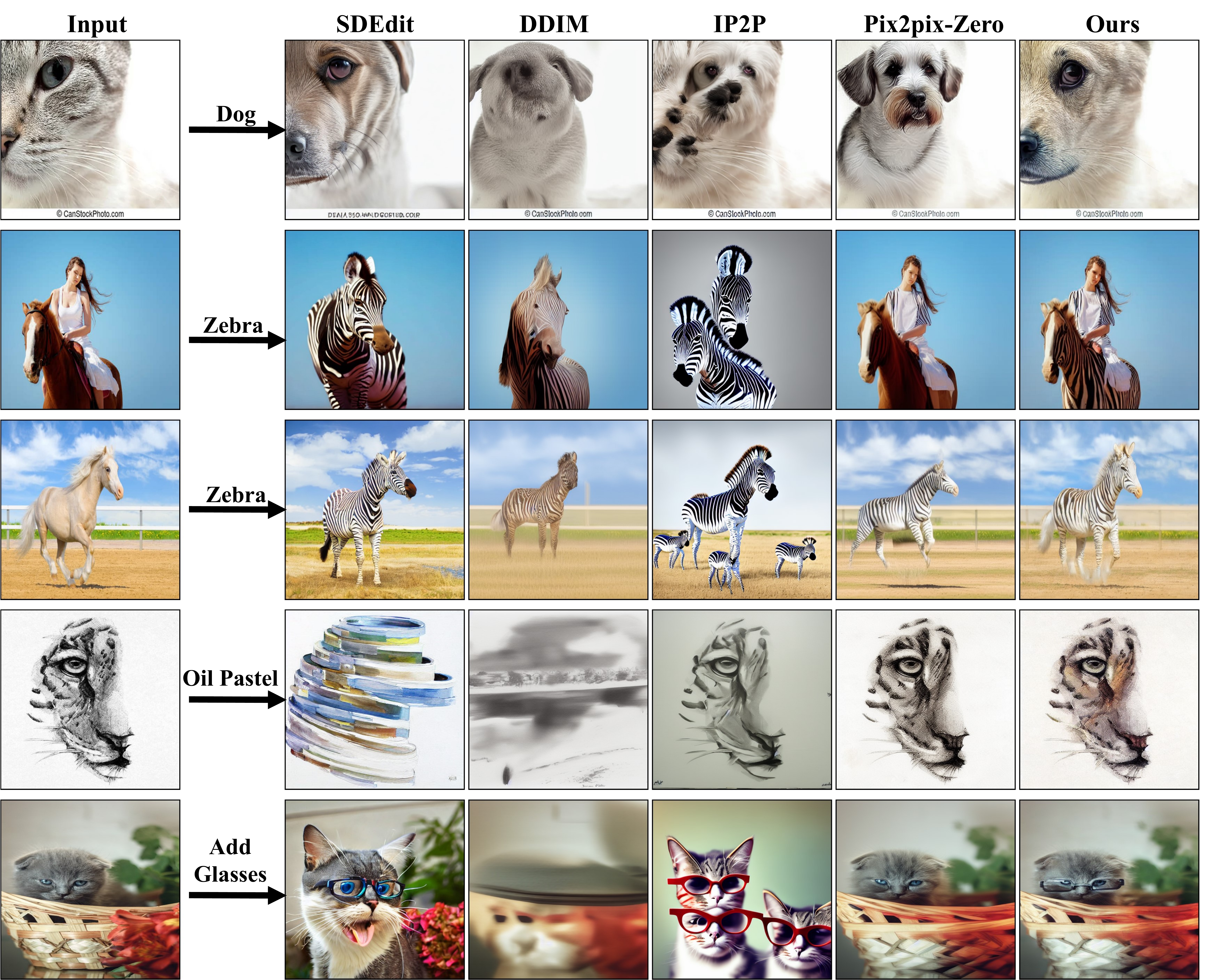}
\caption{Comparisons with different baselines. The SDEdit and DDIM deviated from the original structure, whereas InstructPix2Pix and Pix2Pix-Zero faced challenges during editing. Conversely, our approach successfully preserved the structure of the input image during the editing phase, as depicted in the last column of the image.}
\label{fig:exp_res}
\end{figure*}
\begin{table*}[h]
\centering
\begin{tabular}{ccccccccc}
\hline
                               & \multicolumn{2}{c}{C2D-F}                                  & \multicolumn{2}{c}{H2Z-F}                                  & \multicolumn{2}{c}{C2G-F}                                           & \multicolumn{2}{c}{S2O-F}                                      \\ \cmidrule(lr){2-3} \cmidrule(lr){4-5}  \cmidrule(lr){6-7} \cmidrule(lr){8-9} 
\multirow{-2}{*}{METHOD}        & Clip$\uparrow$                        & Structure$\downarrow$                  & Clip$\uparrow$                         & Structure$\downarrow$                  & Clip$\uparrow$                        & Structure$\downarrow$                  & Clip$\uparrow$                         & Structure$\downarrow$                  \\ \hline
\multicolumn{1}{c}{SDEdit}     & 66.9                               & 0.147                                 & {\color{blue} \textbf{78.7}}                                 & 0.223                                & 76.9                                & 0.133                                 & 56.1                                  & 0.133                                 \\ 

\multicolumn{1}{c}{DDIM}       & 60.2                                & 0.127                                 & 72.5                                 & 0.159                                 & 68.8                                & 0.114                                  & 53.6                                 & 0.122                                 \\ 
\multicolumn{1}{c}{IP2P}       & 72.5                                &  {0.086} & 76.5                                 & 0.256                                 & 74.3                                & 0.155                                 & 66.3                                 & 0.13                                   \\ 
\multicolumn{1}{c}{Pix2Pix-Zero}    & {\color{blue} \textbf{75.2}} & {\color{blue}\textbf{0.071}}                                 & 78.3 & {\color{blue} \textbf{0.106}} & {\color{blue} \textbf{80.3}} & {\color{blue} \textbf{0.047}} & {\color{blue} \textbf{70.7}} & {\color{blue} \textbf{0.060}} \\ \hline
\multicolumn{1}{c}{Ours } & {\color{red} \textbf{75.9}} & {\color{red} \textbf{0.067}} & {\color{red} \textbf{79.6}}  & {\color{red} \textbf{0.094}} & {\color{red} \textbf{81.1}}  & {\color{red} \textbf{0.042}} & {\color{red} \textbf{71.9}} & {\color{red} \textbf{0.051}} \\ \hline
\end{tabular}
\caption{Comparison to baselines. Our method was evaluated with previous image editing methods that using diffusion method by using two evaluation metrics, namely CLIP-Acc and Structure Dist. We performed four different tasks to evaluate the extent of the editing and the degree of structural change implemented. The evaluation results demonstrated that our approach achieved superior CLIP classification accuracy and the lowest Structure Dist score, indicating that our method effectively represents the editing direction while preserving the intricate details of the input image.}
\label{tab:exp_res}
\end{table*}

\section{Experiment}
Our image-to-image translation technique presents a versatile solution for editing real images. In this sections, we showcase the effectiveness of our approach through a series of experiments, utilizing a pre-trained stable diffusion v1.4~\cite{rombach2021highresolution} model.
\subsection{Evaluation}
\textbf{Dataset.} We collected three datasets containing images of CAT, HORSE, and SKETCH from LAION-5B, using an aesthetic filter with a weight of 9 during data collection~\cite{beaumont-2022-clip-retrieval}. Since our method utilizes pre-trained stable diffusion for zero-shot image translation, training is not required, and only 250 images are needed for each dataset for testing purposes. Notably, to enhance the diversity of the sketch dataset, we incorporated data from sketch portraits, sketch animals, sketch cartoon characters, and sketch mythical animals.

\textbf{Task.} Based on the above three datasets, we propose four image-to-image translation tasks for quantitative evaluation:
\begin{itemize}
    \item Cat $\rightarrow$ Dog-free (C2D-F): Achieving a complete transformation from cat to dog while preserving the original appearance.
    \item Cat with Glasses-free (C2G-F): Successfully identifying the eye part of a cat and adding glasses to the cat.
    \item Horse $\rightarrow$ Zebra-free (H2Z-F): Completing the transformation from horse to zebra while maintaining the original look.
    \item Sketch$\rightarrow$ Oil-free (S2O-F): Coloring sketch images in the SKETCH dataset into high-quality oil paintings.
\end{itemize}

\textbf{Metric.} For quantitative evaluations, we utilize two criteria: (1) Successful Application of Edit: we measure whether the desired edit was successfully applied to the input image. (2) Preservation of Input Image Structure: we evaluate whether the structure of the input image is retained in the edited image. We employed CLIP-Acc~\cite{hessel2021clipscore} to calculate the similarity between the edited image and the target text via CLIP. Additionally, we employed Structure Dist~\cite{tumanyan2022splicing} to measure the structural consistency of the edited image.
These quantitative evaluations help in objectively assessing the performance of our method in terms of achieving the intended edits while preserving the structure of the input image during image-to-image translation tasks.

\subsection{ Implementation details} For all tasks of our method, we use 60 steps for DDIM inversion and 60 steps for both reconstruction and editing. In the editing process, we use the Cooperative Update strategy only in the 10th, 15th, 20th, and 25th time steps, and $ \lambda_{rev} $ is scaled down by a certain percentage. In addition, we use a classifier-free guide~\cite{ho2022classifier} for all editing results.

\textbf{Comparisons.} We compare our method with some previous and concurrent diffusion based image editing methods. For a fair comparison, all methods use pretrained stable diffusion, and have the same sampling step. We have chosen four recent works as the baseline for the comparison: (1) SDEdit~\cite{meng2021sdedit} + target domain prompt, (2) DDIM~\cite{song2020denoising} + target domain prompt. (3) InstructPix2pix~\cite{brooks2022instructpix2pix} + target domain prompt(IP2P). (4) Pix2Pix-Zero~\cite{parmar2023zero}. SDEdit uses the hugging face's StableDiffusion Img2Img API, which uses the diffusion-denoising mechanism, and the other baselines use the official implementation.

\subsection{Experiment Result}

In Fig.~\ref{fig:exp_res}, we compare our method with the baseline approaches. It can be observed that SDEDIT + target domain prompt and DDIM + target domain prompt were successful in converting the images into transformed targets, but faced challenges in preserving the input image structure while editing. Although Instruct Pix2Pix can preserve the structure through the use of the original image's cross-attention map as a hard constraint, they are not able to achieve the desired editing results. Compared to Instruct-Pix2Pix, Pix2Pix-Zero preserves the overall appearance better, but it still falls short of achieving the desired editing effects. In contrast, our method is able to preserve the shape and achieve the desired editing effect that other methods fail to accomplish.

In Table ~\ref{tab:exp_res}, we have compared our method to the baseline, with red indicating the best results and blue indicating the second best. Our method exhibits a higher CLIP-Acc, indicating that we are able to achieve accurate and meaningful image editing while preserving the structure and background of the original input image with minimal structure errors.

\subsection{Ablation Study}
\begin{figure}[t]
\centering
\includegraphics[width=0.89\linewidth]{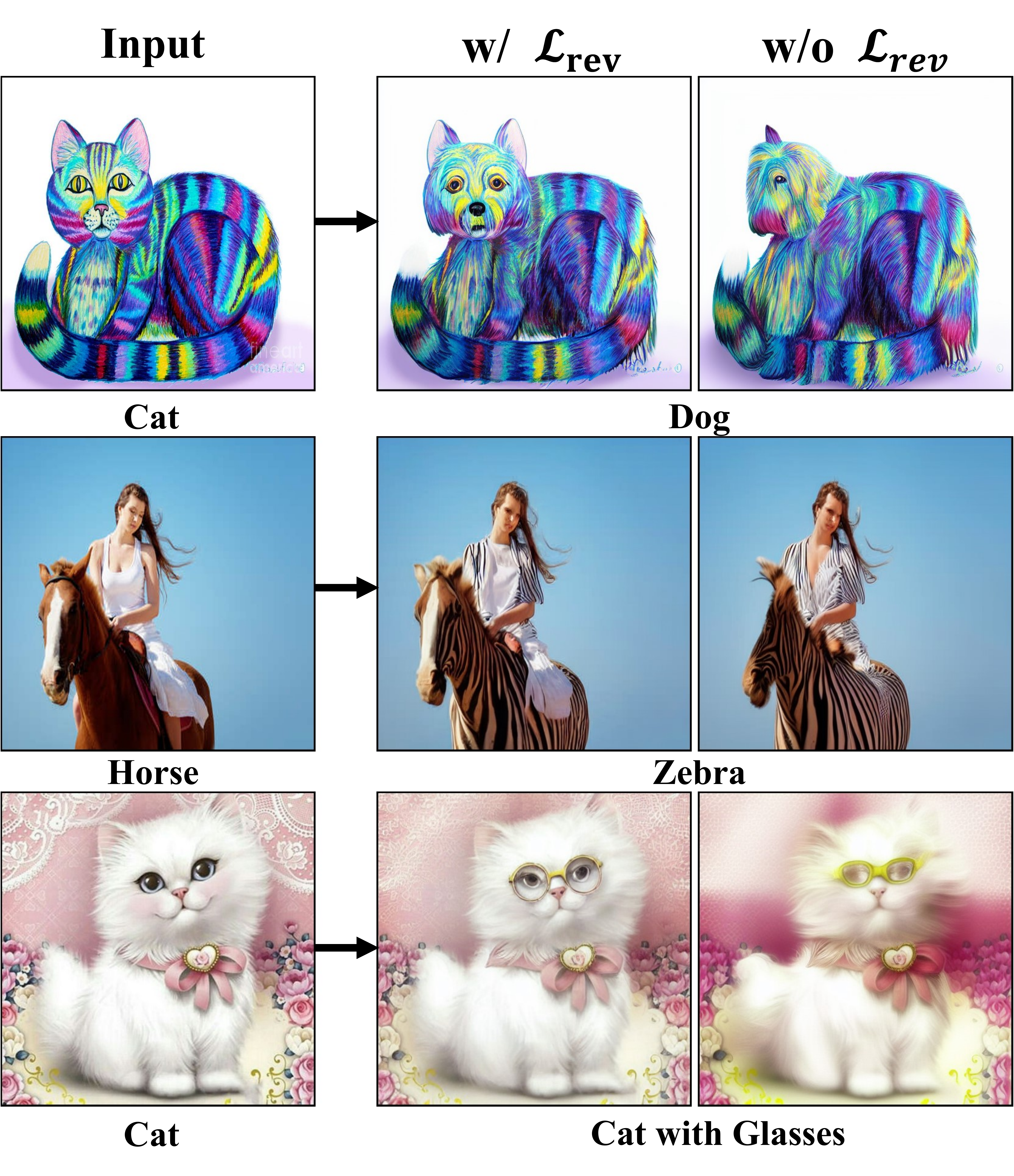}
\caption{Effectiveness of Cooperative Update on structure preservation. With $\mathcal{L}_{{rev}}$, the structure is well-preserved for objects.}
\label{fig:AB}
\end{figure}

\begin{table}[t]
\footnotesize
\centering
\begin{tabular}{ccccc}
\hline
                         &                             &                          & \multicolumn{2}{c}{H2Z-F}                                                    \\ \cmidrule(lr){4-5} 
\multirow{-2}{*}{Config} & \multirow{-2}{*}{Reference} & \multirow{-2}{*}{Fusion} & Clip$\uparrow$                         & Structure$\downarrow$                  \\ \hline
A                        & \makecell{Initial Prompts}         & \textbf{$\overline{\ \ \ \ }$}                        & 78.3                                 & 0.106                                 \\ \hline
B                        & \makecell{ReGen with \\ Rich Prompts}                & Simple            & 78.8                                 & 0.114                                 \\ \hline 
C                        & \makecell{ReGen with \\ Rich Prompts}                   & Sliding              & {\color{blue} \textbf{78.9}} & {\color{blue} \textbf{0.095}} \\ \hline
D(Ours)                        & \makecell{ReGen with \\ Rich Prompts w/ $\mathcal{L}_{{rev}}$}                 & Sliding             & 
{\color{red} \textbf{79.6}} & {\color{red} \textbf{0.094}} \\ \hline
\end{tabular}
\caption{Ablation study. We performed an ablation study where we added different components of our method one at a time and observed their effects. Config A represents Pix2Pix-Zero. Config B and C illustrate that ReGeneration Learning with rich prompts and sliding fusion. Config D demonstrates how effectively the structure can be maintained using Cooperative Update.}
\label{tab:AB_res}
\end{table}

In our ablation experiments, we conducted a thorough analysis of each component to evaluate its effectiveness. The results of these ablation experiments are summarised in Table ~\ref{tab:AB_res}, which compares the performance of four different configurations. Config A is represented as the original configuration of Pix2Pix-Zero, which uses the cross-attention map of the initial prompt as the reference for the editing phase. In contrast to employing a solitary reconstructed cross-attention map as a reference in the editing phase, Config B fuses the cross-attention maps of rich prompts at each time step as a reference during editing phase. The experimental results demonstrate that this approach yields enhancements in CLIP-Acc but is accompanied by a reduction in structure preservation. This suggests that this approach encourages expression of the target domain, but has some drawbacks in maintaining appearance. Rather than simply fusing rich prompts, Config C further uses the sliding fusion, which combines pre- and post-moment features of rich prompts. This method effectively enhances both CLIP-Acc and structural preservation.

Finally, Config D introduces Cooperative Update, which further improves the CLIP-Acc and structural performance. The impact of Cooperative Update is depicted qualitatively by comparing Config C, and Config D in Fig.~\ref{fig:AB}. When editing images with unusual perspectives, Config C maintains appearance, but the image quality is poor, while the Config D outperforms the other configurations, retaining appearance fidelity and completing the edit successfully. For example, as illustrated in Fig.~\ref{fig:AB}, it can be observed that the edited image fails to maintain the shape of the original image when Cooperative Update is not used.
\begin{figure}[t]
\centering
\includegraphics[width=0.89\linewidth]{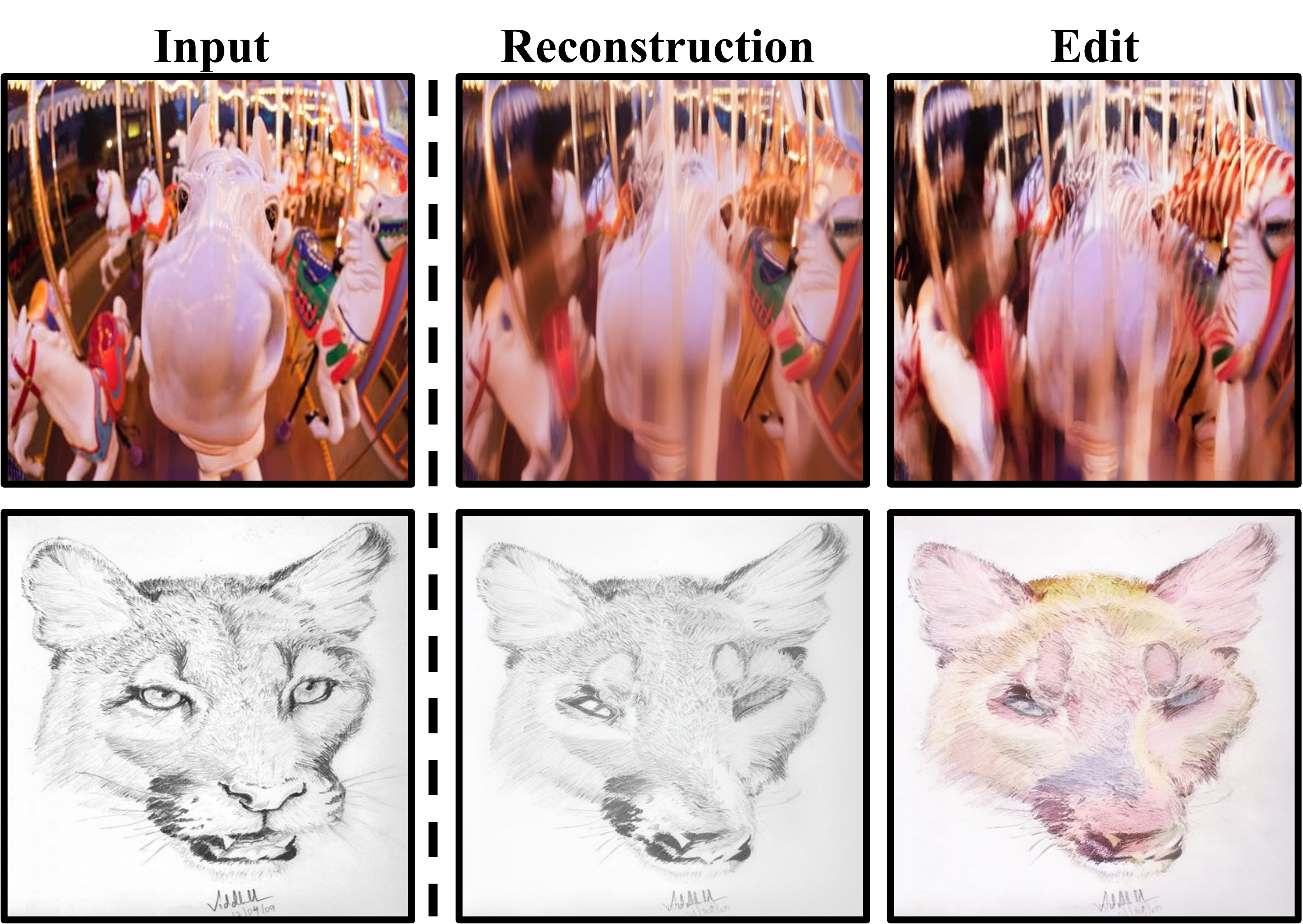}
\caption{Limitation. Suppose there are significant differences between the reconstructed results and the original image, our method may fail to achieve the intended editing results.}
\label{fig:limitation}
\end{figure}
\section{Conclusion and Limitations}
In this paper, we propose a novel ReDiffuser for zero-shot image-to-image translation. In the reconstruction phase, we produced cross-attention maps of rich prompts based on Regeneration Learning and employed the sliding prompt fusion approach to merge these prompts as a reference to guide the image editing. During the editing phase, we employed the reference map as a guide to preserve the structure of the original image and encourage the representation of the target domain information. Furthermore, we utilized a strategy called Cooperative Update to ensure consistency of appearance. Compared with the previous diffusion-edit methods, our proposed ReDiffuser is prompt-free and structure-preserving. 

Similar to Pix2Pix-Zero, our image editing approach bases on the result of the DDIM inversion. As shown in Fig.~\ref{fig:limitation}, if the reconstructed image quality is poor, it may adversely impact our editing results. Therefore, addressing the issue of poor reconstruction results will be a key focus of our future research.

{\small
\bibliographystyle{ieee_fullname}
\bibliography{RebuttalTemplate}
}

\end{document}